\documentclass[twocolumn,letterpaper]{IEEEAerospaceCLS}  

\usepackage[]{graphicx}    
\newcommand{\ignore}[1]{}  
\usepackage{hyperref}
\usepackage{tabularx,booktabs}
\usepackage{subcaption}
\usepackage[noadjust]{cite}
\usepackage{url}
\usepackage{dblfloatfix}    

\usepackage{lipsum}
\usepackage{fancyhdr}
\usepackage{amsmath,amssymb}
\usepackage{gensymb}

\usepackage{fancyhdr}
\usepackage{lipsum}

\begin{document}
\title{Seasonal Station-Keeping of Short Duration High Altitude Balloons using Deep Reinforcement Learning}

\author{%
Tristan K. Schuler\\ 
U.S. Naval Research Laboratory\\
4555 Overlook Ave. S.W\\
Washington, D.C. 20375\\
tristan.k.schuler.civ@us.navy.mil
\and 
Chinthan Prasad\\
U.S. Naval Research Laboratory\\
4555 Overlook Ave. S.W\\
Washington, D.C. 20375\\
chinthan.b.prasad.civ@us.navy.mil
\and 
Georgiy Kiselev\\
U.S. Naval Research Laboratory (NREIP Intern)\\
4555 Overlook Ave. S.W\\
Washington, D.C. 20375\\
georgiy.a.kiselev.ctr@us.navy.mil
\and 
Donald Sofge\\ 
U.S. Naval Research Laboratory\\
4555 Overlook Ave. S.W\\
Washington, D.C. 20375\\
donald.a.sofge.civ@us.navy.mil
\thanks{U.S. Government work not protected by U.S. copyright}              
}

\maketitle

\fancypagestyle{firstpage}{
    \fancyhf{} 
    \fancyfoot[C]{DISTRIBUTION STATEMENT A: Approved for public release, distribution is unlimited.} 
    \renewcommand{\headrulewidth}{0pt}
}
\pagestyle{plain}

\maketitle
\thispagestyle{firstpage}
\pagestyle{plain}

\begin{abstract} 
Station-Keeping short-duration high-altitude balloons (HABs) in a region of interest is a challenging path-planning problem due to partially observable, complex, and dynamic wind flows. Deep reinforcement learning is a popular strategy for solving the station-keeping problem. A custom simulation environment was developed to train and evaluate Deep Q-Learning (DQN) for short-duration HAB agents in the simulation. To train the agents on realistic winds, synthetic wind forecasts were generated from aggregated historical radiosonde data to apply horizontal kinematics to simulated agents. The synthetic forecasts were closely correlated with ECWMF ERA5 Reanalysis forecasts, providing a realistic simulated wind field and seasonal and altitudinal variances between the wind models. DQN HAB agents were then trained and evaluated across different seasonal months. To highlight differences and trends in months with vastly different wind fields, a Forecast Score algorithm was introduced to independently classify forecasts based on wind diversity, and trends between station-keeping success and the Forecast Score were evaluated across all seasons.
\end{abstract}

\tableofcontents

\section{Introduction}\label{section:Introduction}

High-altitude balloons (HABs) with altitude control capabilities can leverage opposing winds at different altitudes to achieve limited horizontal control.  By adjusting altitude to ``ride'' favorable winds back and forth to maintain position over an area of interest, HABs can perform station-keeping maneuvers, or with higher wind diversity, conduct more complicated trajectories. Path planning for HABs is non-trivial due to complex dynamic wind fields and partial observability of forecasts.  Wind flows constantly change in time, position, and altitude.  Some wind trends lead to more favorable and predictable station-keeping performance while some wind configurations render station-keeping infeasible.  Developing robust path-planning algorithms that can handle forecast errors also requires wind models that realistically deviate from forecasts.

Deep Reinforcement Learning (DRL) is a subset of reinforcement learning that incorporates neural networks in traditional reinforcement learning. At a high level, DRL can be used to train agents to engage in optimal behaviors in the presence of complex environments without explicit modeling of the state-action transition matrix. To determine these policies, agents are typically trained in simulation for thousands to millions of ``episodes'' depending on the problem complexity. In recent years, several state-of-the-art DRL algorithms have emerged and are actively utilized in the scientific community, such as Deep Q-Networks (DQN), Proximal Policy Optimization (PPO), and actor-critic methods such as A2C and SAC. In this work we apply DQN, a popular state-of-the-art off-policy DRL algorithm,  for completing objectives in complex environments based on past experiences. DQN was developed by Deepmind in 2013 where they successfully demonstrated beating human experts at playing Atari 2600 games \cite{mnih2013playing}. 

In traditional Q-Learning, the Bellman Equation, shown in Eq. \ref{Bellman}, is used to approximate the value function $Q(s, a)$ of the state $s$ and the action $a$ at the Markovian decision point. The value function $Q(s, a)$ represents the inherent value, looking forward in time, that a particular decision (state, action pair) can have. 

\begin{equation}
    Q(s,a) = r + \gamma max_{a'}Q(s',a')
    \label{Bellman}
\end{equation}

In situations where the state and action space are discrete, tabular representations of the value function are the most efficient. However, in situations with complex environments, high state dimensionality, and/or continuous state spaces, estimating tabular values for the value function is not feasible. DQN employs a neural network to approximate the value function and results in a more efficient and robust policy. Since the original publication, DQN and it's many variations have been implemented on autonomous and robotic systems. DQN has been implemented in many real-world complex path planning situations such as in a warehouse dispatching system \cite{yang2020multi}, with coastal ships \cite{guo2021path}, autonomous driving \cite{chen2020conditional} and drones \cite{azar2021drone}. 


\subsection{Related Work}

Google Loon provided the first example of using Deep Reinforcement Learning to successfully station-keep super-pressure balloons in real-time, both in simulation and flight tested on HABs in the Pacific Ocean.  They trained HAB agents in a custom simulation environment using  Distributional Reinforcement Learning with Quantile Regression (QR-DQN) and achieved an average station-keeping time within a region of 50 km (TWR50) of approximately 55\% over multi-day, sometimes multi-week flights \cite{bellemare2020autonomous}. 
Brown et al. showed through simulation with ERA5 wind data and tree search path planning, how autonomous HABs leveraging winds in the
stratosphere can have significantly different success rates dependent on season and geographic region of operation \cite{brown2024seasonal}.
Similarly,  Schuler et. al also investigated wind diversity trends in the Western Hemisphere from over 1 million aggregated radiosonde data points over a decade with respect to controllable HABS \cite{schuler2025wind}.
Jeger et al. used the Soft Actor-Critic algorithm to successfully control short duration balloons from a start point to a predefined target in under 1.5 hours and demonstrated the algorithm on indoor and outdoor prototypes \cite{jeger2023reinforcement}. Xu et al. provide a detailed dynamic model for air-ballasted altitude control balloons and demonstrated station keeping using a Double Dueling Deep Q-Learning (D3QN) framework with priority experience replay based on high-value samples (HVS-PER) \cite{xu2022station}.
Saunders et al. created a custom simulation environment to train short-duration resource constrained weather balloons with a vent and ballast system. Using the Soft Actor-Critic (SAC) algorithm, they trained simulated HABs to station keep within TWR50 of 25\% of the flight using ECWMF wind data \cite{saunders2023}. 
Similarly,  Gannetti et al. simulated sounding balloons with an altitude control system using DQN on ERA5 winds and achieved an average simulated TWR50 of 12.5\% \cite{gannetti2023navigation}.



\section{Station Keeping with DQN}\label{section:SimEnvforDQN}

To successfully perform station-keeping, HABs typically oscillate between altitudes with opposing winds, stay in calm wind regions ($<$ 2m/s) regions, or some combination of the two. Most altitude-controlled HAB platforms operate between 18-28 km, above the typical max altitude regulation for many countries (such as FL600 for the USA).  Luckily, this region of the atmosphere frequently has vertical wind diversity for maneuvering HABs, where different altitudes can have opposing winds. The probability of available diverse winds in a region of interest substantially differs geographically and seasonally. 

Throughout a flight, the HAB can adjust altitude any number of times, as well as leave and re-enter the region any number of times, resulting in many unique trajectories and solutions to the problem. We use deep reinforcement learning to approach the station-keeping problem as opposed to a more deterministic controller because real-world wind flows are extremely complex and dynamic, as well as often heavily differing from weather forecasts.  For a reinforcement learning approach to station keeping, we design the simulation environment in a game-like structure to learn a policy, introducing simple actions, observation states, and positive rewards for making good actions towards the station-keeping goal.  We implement the popular DQN deep reinforcement learning algorithm to train a policy that maximizes rewards based on state action pairs. Training the DQN policy involves running hundreds of thousands of individual simulations, further referred to as episodes, updating the neural network as new conditions are encountered, and building off of past experiences to determine optimal actions for any given state. For evaluating trained agents, we use a time within the region of 50 km (TWR50) metric to measure station-keeping success.


\subsection{Simulation Setup}\label{section:SimEnvforDQN_SimStructure}

\begin{figure*}[t]
    \centering 
    \includegraphics[width=\textwidth]{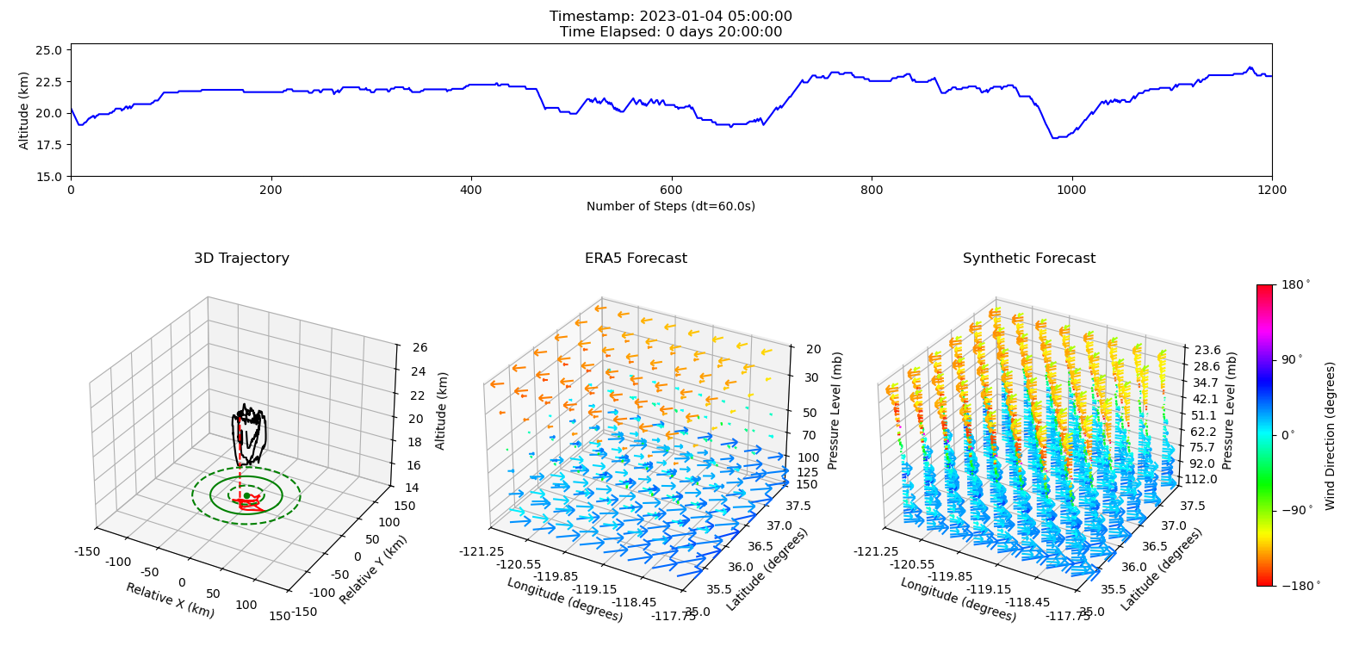} 
    \caption{Altitude Profile and 3D trajectory of a trained DQN HAB agent in simulation with ERA5 forecasts as observation, and synthetic winds for movement.}
    \label{fig:simulator} 
\end{figure*}

In the simulation, we use $u$ and $v$ wind data from ERA5 Reanalysis as our forecast, and pre-generated synthetic winds (discussed in Section \ref{section:SynthForecastModeling}) as our ground truth for dictating HAB movement. ECWMF ERA5 Reanalysis forecasts provide a large collection of historical atmospheric variables through reanalysis, including wind vectors at 137 static pressure levels at a maximum horizontal resolution of 0.25\degree \cite{hersbach2018era5}. For operational balloon missions ERA5 would be replaced by a forward-looking forecast and Synthetic winds would be the true unknown winds. At the start of an episode, we prepare a HAB arena (150 x 150 x 10 km) with a subset of 3D flow fields derived from a random coordinate within the bounds of a pre-downloaded and processed master forecast (e.g., Southwestern United States).  The HAB is then initialized to a random altitude at the station coordinate. We standardized the episode length to be 20 hours long at a resolution of 1 minute. 
At each simulation step, the horizontal flow is applied according to the $u$ and $v$ wind components from the synthetic winds forecast, while the observation space is updated with relative wind profiles from the ERA5 forecast. 
Throughout an episode, intermediate rewards are appended to an overall score based on the current observation after executing an action; the DQN policy aims to maximize the overall reward by training the network to choose the best intermediate actions.
 Figure \ref{fig:simulator} shows a snapshot of a trained DQN agent station-keeping in the simulation environment GUI.  The simulator GUI shows the 3D trajectory, altitude profile, and corresponding ERA5 and synthetic forecasts for the simulated coordinates and timestamp.


\subsection{Solar Balloon Dynamics}\label{section:SimEnvforDQN_SolarBalloonDynamics}

For training short-duration HABs with DQN, we selected Vented Solar High Altitude Balloons (SHAB-Vs) for simulating HAB dynamics. Solar balloons utilize a lightweight material that absorbs solar radiation, providing a free source of lift and eliminating the need for lighter than air gas or heat source. Solar balloons can fly as long as there is unobstructed sunlight; 12 hours in equatorial regions but up to 16+ hours at higher latitudes in the summer months. Researchers have logged hundreds of flight hours with the original heliotrope design as well as modified versions, lifting up to 3 kg payloads and conducting multi-hour stratospheric research \cite{bowman2020multihour, schuler2022solar, swaim2024performance, Lien2024EarthSHAB}.  Solar balloons have additionally been proposed as a feasible platform for exploring Venus' atmosphere due to the thick, highly reflective cloud layer \cite{schuler2022long}. Recent development of a novel mechanical vent for solar balloons allows SHAB-Vs to adjust altitude precisely between a typical altitude range of 15-23 km and demonstrated initial manual station-keeping in the Southwestern United States \cite{schuler2023altitude}.



 \subsection{Action Space}

At each simulation step, the SHAB-V agent can perform three discrete actions: ascend, maintain altitude, or descend. In heliotrope solar balloon flight experiments conducted in July-September 2021, Swain et al. showed that the average ascent and descent rates of non-vented heliotrope balloons are 1.8 $\pm$ 0.14 and 2.8 $\pm$ 0.3 m/s respectively \cite{swain2024heliotrope}. Additionally, SHAB-V flight experiments conducted by Schuler et al. implemented an altitude controller with standard deviation $\sigma$ = 1.25 m/s. We simulate the 3 discrete actions and reasonable errors by mapping each venting action to the normal distributions found in Table~\ref{tab:actionspace}. These distributions holistically represent the motion profile experienced via the onboard altitude controller.


\begin{table}[h]
\renewcommand{\arraystretch}{1.2}
\caption{Action Space for Training}
\centering
\begin{tabular}{|c|c|c|}
\hline
\bfseries Action & \bfseries Sampling Distribution \\
\hline
Ascend & $1.80 \pm 0.14$ m/s\\
Descend & $2.80 \pm 0.30$ m/s\\
Stay & $0.00 \pm 1.25$ m/s\\
\hline
\end{tabular}
\label{tab:actionspace}
\end{table}

Yajima et al. showed that the lateral motion of HABs is a function of the relative velocity between the balloon and the wind and dictated by the composition of the balloon, payload mass, and varying atmospheric properties at different altitudes \cite{yajima2009ballooning}. Smaller radiosonde balloons (of mass 0.3 kg to 2 kg) require between 20 to 40 seconds to respond to a 2 m/s velocity step change \cite{yajima2009ballooning}. Over large distances and multi-hour timescales, this delay in matching lateral wind velocities creates small deviations when considering station-keeping trajectories that are constantly adapting to wind flow changes. Additionally, winds typically experience a gradual shift between altitudes of different directional flows, with turbulent regions occurring less often. With these conditions, we assume the lightweight SHAB-Vs match the wind instantaneously in our simulation environment. Horizontal wind velocities from pre-computed synthetic forecasts, discussed in Section \ref{section:SynthForecastModeling}, are applied to the SHAB-V over the course of a 60-second time step ($\delta t$). A 60-second time step allows for reasonable simulation training time and also renders the time required to reach zero relative velocity negligible.



\subsection{Observation Space}
\begin{table}[h]
\renewcommand{\arraystretch}{1.2}
\caption{Observation Space for Controller Input}
\centering
\begin{tabular}{|c|c|}
\hline
\bfseries Observation & \bfseries Range\\
\hline
HAB Altitude & $15 - 25$ km\\
Relative Distance & $0 - \infty$ km\\
Relative Bearing & $0 - 180\degree$\\
Wind Column & \{Altitude, Magnitude, \\
            & Relative Bearing\} * N \\
\hline
\end{tabular}
\label{tab:observationspace}
\end{table}

The observation space for training the DQN policy is a subset of the agent's full state and is shown in Table \ref{tab:observationspace}. This observation space does not include all simulator state information such as velocity, coordinates, or an entire forecast as DRL algorithms may take longer to train as a function of a more complex observation space. By converting coordinates to relative distance and bearing, we are able to reduce the dimensionality of the observation vector.
The relative distance and relative bearing are from the balloon to the central coordinate for station-keeping where the maximum relative bearing is 180\degree and would be a HAB traveling on a trajectory directly away from the central coordinate. The wind column is a triple that encodes relative wind velocities at the current coordinate to the central coordinate from an ERA5 forecast along with the true altitude of those coordinates. The number of levels included in the wind column remains static for the entire training set (for the altitude region of 15-25 km we include 7 levels between 20 and 150 hPa). Because ERA5 is a pressure-based forecast and we extrapolate the altitude values from the geopotential variable, altitude is not constant across a forecast and why we include it in the wind column triple.  For small forecast areas (only a few degrees of latitude and longitude across) the variance in altitude at the same pressure level is typically within $\pm$ 100 m.  By including true ERA5 altitudes in the wind column, the DRL algorithm is able to implicitly learn discrepancies between ERA5 and the synthetic winds in different seasons.
In the future, we plan to experiment with expanding the number of altitude levels in the wind column vector based on a combination of ERA5 forecasts and previous actions, experimenting with relative altitude values, and including uncertainty estimates.


\subsection{Reward Functions}\label{section:SimEnvforDQN_RewardFunctions}

For short-duration station-keeping, we experimented with the distance-based reward proposed by Bellemare et al. in Eq. \ref{eq:google} as well as some variations \cite{bellemare2020autonomous}. For short-duration station-keeping missions we had the best performance with the piecewise reward function that introduces an inner radius as shown in Eq. \ref{piecewise_reard}.  For short-duration HAB station-keeping, if the winds have good station-keeping probabilities, they frequently do throughout the entire episode, whereas multi-day or multi-week flights can experience significantly different wind profiles over a longer time scale. The Euclidean Reward function in Eq. \ref{euclidan_reard} performed slightly worse than the piecewise reward function and about the same as Google's reward function in our simulation environment, and we believe would be the best candidate for navigating to waypoints as opposed to assuming the HAB is already in the station-keeping region. 

With SHAB-Vs, we ignore power as a resource constraint because the HABs can lift several kilogram payloads and can carry enough payload to continuously vent. Similarly, longer duration altitude controlled balloon platforms such as air ballasted super pressure HABs and vent and ballasted zero pressure HABs could also reduce or eliminate power constraints for short duration flights.  Unlike other HAB platforms, SHAB-Vs do not have a lighter-than-air gas as a resource constraint because the lift is generated via heating ambient atmospheric air.  Therefore, the altitude profile and horizontal motion are the most important dynamics to replicate.

Google Loon Reward Function:

\begin{equation}
r_{\mathrm{dist}}(\Delta) =
    \begin{cases}
        1.0 & \text{if } \Delta < \rho_{50km} \\
        c_{\mathrm{cliff}}2^{-(\Delta-\rho)/\tau)} & \text{otherwise }
    \end{cases}
    \label{eq:google}
\end{equation}

Piecewise Reward Function:

\begin{equation}
r_{\mathrm{dist}}(\Delta) =
    \begin{cases}
        2.0 & \text{if } \Delta \leq \rho_{25km} \\
        1.0 & \text{if }  \rho_{25km} < \Delta \leq \rho_{50km} \\
        c_{\mathrm{cliff}}2^{-(\Delta-\rho)/\tau)} & \text{otherwise }
    \end{cases}
    \label{piecewise_reard}
\end{equation}

Euclidean Reward Function:

\begin{equation}
r_{\mathrm{dist}}(\Delta) =
    \begin{cases}
        |\Delta| & \text{if } \Delta < \rho_{50km} \\
        c_{\mathrm{cliff}}2^{-(\Delta-\rho)/\tau)} & \text{otherwise }
    \end{cases}
    \label{euclidan_reard}
\end{equation}

\subsection{Training and Hyperparameter Tuning}\label{section:SimEnvforDQN_TrainingAndHyperparam}

\begin{figure}[h]
    \centering 
    \includegraphics[width=.5\textwidth]{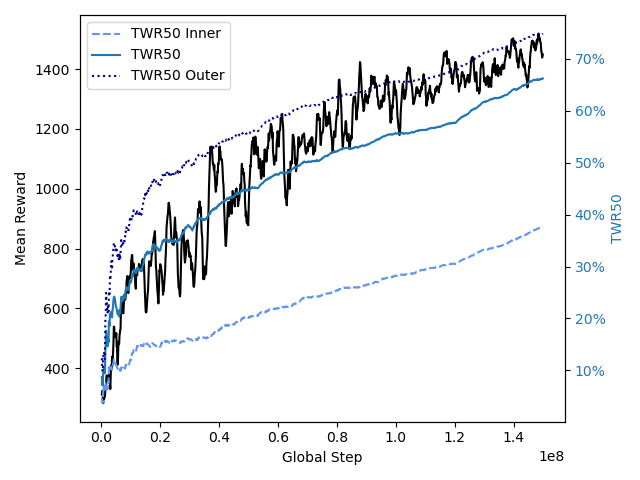}
    \caption{Learning curve over 1.5 million timesteps for a station-keeping DQN HAB agent} 
    \label{fig:Learning} 
\end{figure}

Figure \ref{fig:Learning} shows an example of one of the best-performing trained models for an HAB in the month of April trained for 150 million time steps and reaching an average TWR50 of 75\%. Training was only done on forecasts with decent wind diversity, so the final TWR50 is higher than the overall mean. The TWR50 and TWR75 closely follow the mean reward learning curve whereas the inner TWR25 is much lower. The mean reward started to plateau toward the end of training. However, all 3 of the TWR metrics are still increasing, suggesting longer training times could lead to improved station-keeping results.  

The best-trained models for short-duration HAB station-keeping are extremely sensitive to hyperparameters of the DQN algorithm as well as the geographic area and season where the agent was trained. We used Optuna, an open-source automated hyperparameter tuning framework, to train our initial models \cite{akiba2019optuna}. Overall, the learning rate and exploration rate had the most significant impacts on success. The best-performing models had learning rates between 1e-5 and 1e-4 and a high exploitation to exploration ratio (how often the agent takes random actions as opposed to picking actions from the trained policy.) For the exploration strategy, the best-performing models started with taking random actions approximately 25-50\% of the time at the beginning of training and linearly decayed for about a third of the overall training time to take random actions 10-20\% of the rest of the time, a more deterministic policy. 



From preliminary investigations, we noticed that models trained on one region (e.g., the Southwestern United States) but evaluated on another region (e.g., an equatorial region) performed significantly worse than models trained on the other region.  The primary reason is that the HAB agent is exposed to wind flows never seen before when only training on a medium-sized region. Equatorial regions typically have significantly higher wind diversity in all dimensions (vertically, horizontally, and temporally) than the Southwestern United States, and also more deviations between ERA5 and synthetic forecasts.  Combining models, implementing curriculum learning, and other optimization strategies could help the trained DRL agents be more robust to much larger areas and timescales, but this is currently out of the scope of this paper for short-duration HABs.

\section{Synthetic Forecast Modeling}\label{section:SynthForecastModeling}

To have autonomous HABs that are more robust to inaccurate weather forecasts, realistic deviations from available weather forecasts are necessary.  However, determining and applying these deviations is nontrivial; while adding simple noise to the ERA5 forecasts could be one approach, it is not very realistic for several reasons. Often, real winds at a particular altitude during flight can be up to 180\degree off from the forecast.  Additionally, winds at one altitude level do not necessarily linearly shift in speed and direction when there are significant changes between time steps. 
Finally, and most importantly, winds at one altitude level are frequently decoupled from other altitude levels, hence why opposing winds and wind diversity are possible. 


Simulating realistic winds and uncertainty in forecasts is a major challenge due to a lack of in situ weather data in the upper troposphere and in the lower stratosphere. Popular Reanalysis forecasts such as ECWMF and NASA GEOS only report wind data at specific levels, that lack vertical resolution. NOAA's GFS forecast is similar but forward predicts forecasts rather than reanalysis;  we will most likely use GFS forecasts for real-world SHAB flights with trained models in the future. Since pressure falls exponentially from sea level to the upper atmosphere,  higher altitudes have decreased vertical resolution. Within the typical HAB operating altitude region, most forecasts only report between 5-10 pressure levels of wind data. 


\begin{figure}[!ht]
\centering
    \begin{subfigure}[t]{0.4\textwidth}
    \centering
    \includegraphics[width=\textwidth]{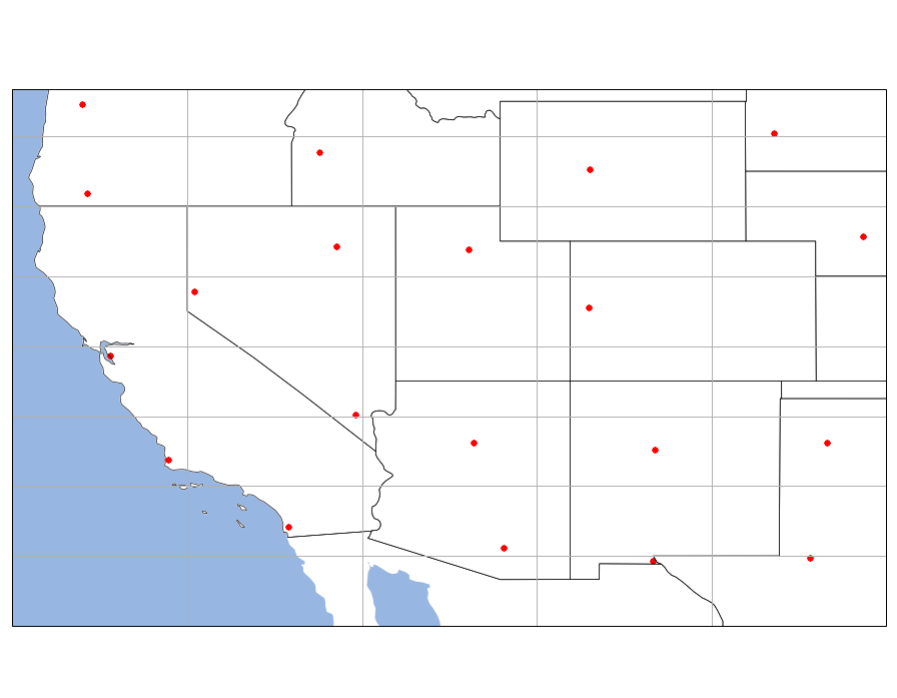}
    \caption{Map of Radiosonde Locations in the Southwestern United States} \label{fig1}
\end{subfigure}\hfill
\begin{subfigure}[t]{0.4\textwidth}
    \centering
    \includegraphics[width=\textwidth]{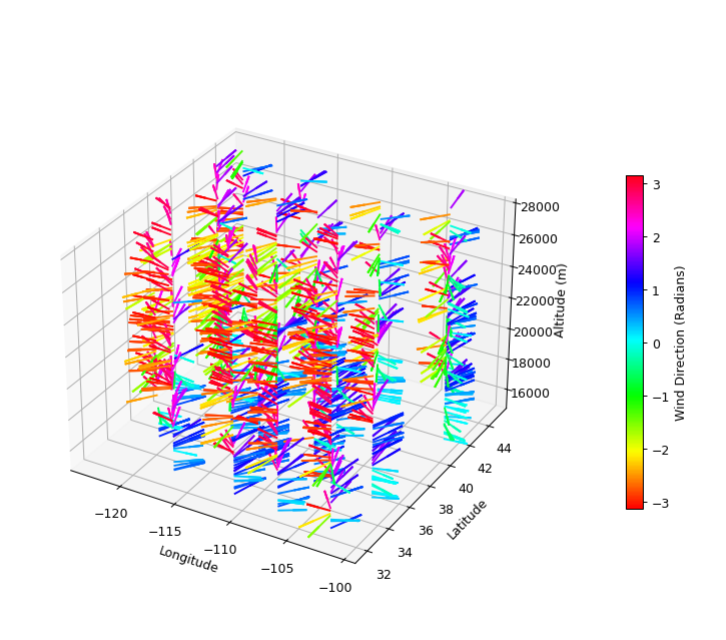}
    \caption{3D wind vector visualization from unprocessed radiosonde measurements} \label{fig2}
\end{subfigure}\hfill
\begin{subfigure}[t]{0.4\textwidth}
    \centering
    \includegraphics[width=\textwidth]{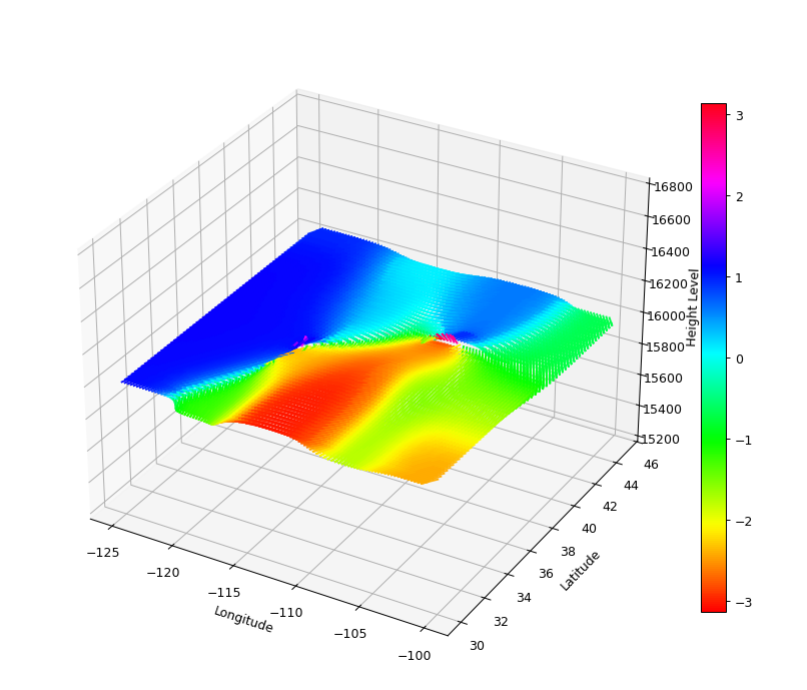}
    \caption{Smoothed Synthetic Winds at altitude level 16 km} \label{fig2}
\end{subfigure}\hfill

\caption{Synthetic Wind Generation from Aggregated and Interpolated Radiosonde Data in the Southwestern United States on August
23, 2023 at 1200 UTC}
\label{fig:synthwinds}
\end{figure}

\begin{figure*}[b] 
    \centering
    \subfloat[ERA5 Model (Jan)]{\includegraphics[width=0.32\textwidth]{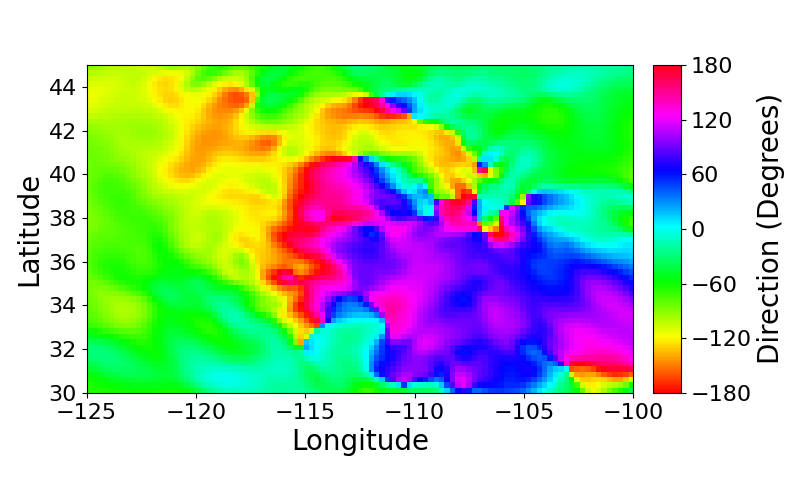}\label{fig:01_17_50_a}}%
    \hspace{0.01\textwidth} 
    \subfloat[Synthetic Model (Jan)]{\includegraphics[width=0.32\textwidth]{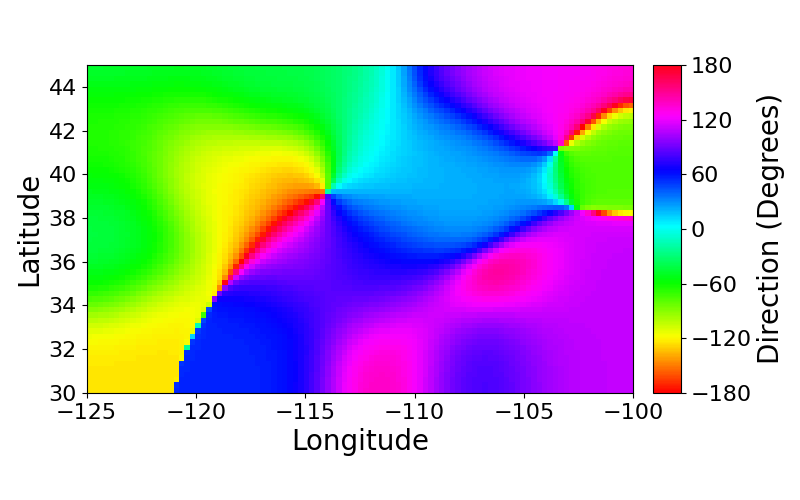}\label{fig:01_17_50_b}}%
    \hspace{0.01\textwidth} 
    \subfloat[Model Difference (Jan)]{\includegraphics[width=0.32\textwidth]{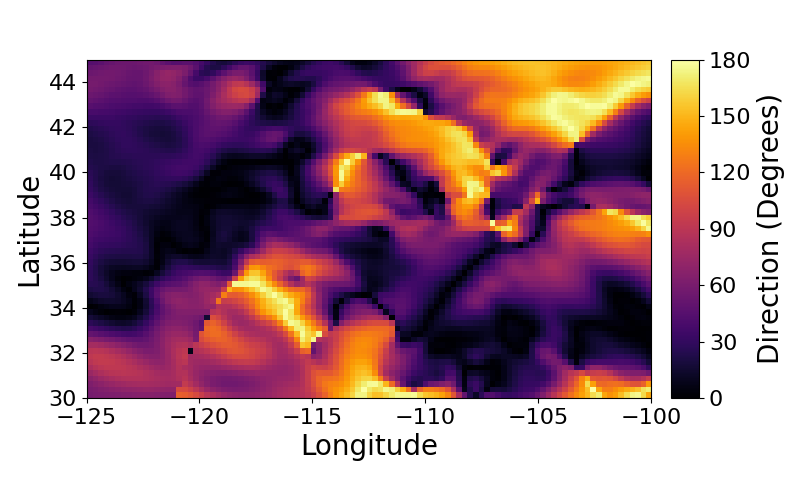}\label{fig:01_17_50_c}}
    
    \vspace{0.1cm} 
    
    \subfloat[ERA5 Model (Jul)]{\includegraphics[width=0.32\textwidth]{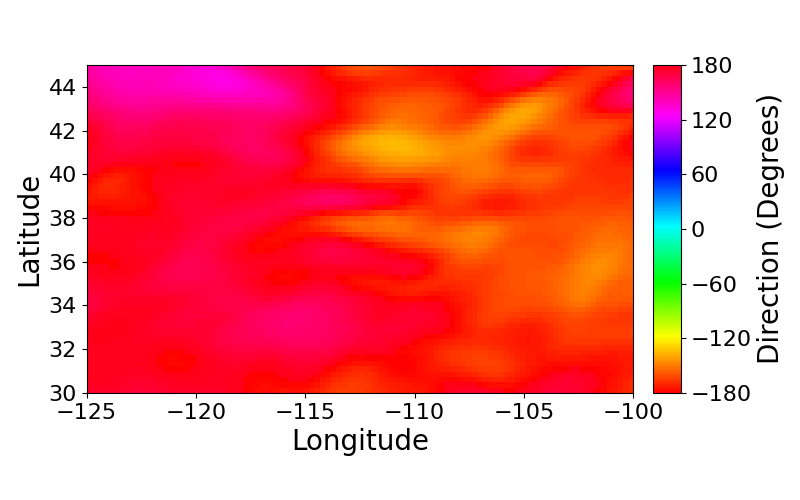}\label{fig:07_17_50_a}}%
    \hspace{0.01\textwidth} 
    \subfloat[Synthetic Model (Jul)]{\includegraphics[width=0.32\textwidth]{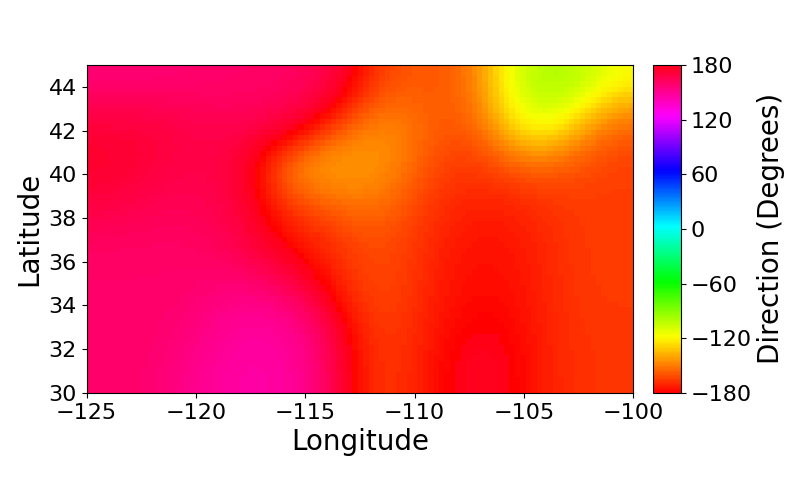}\label{fig:07_17_50_b}}%
    \hspace{0.01\textwidth} 
    \subfloat[Model Difference (Jul)]{\includegraphics[width=0.32\textwidth]{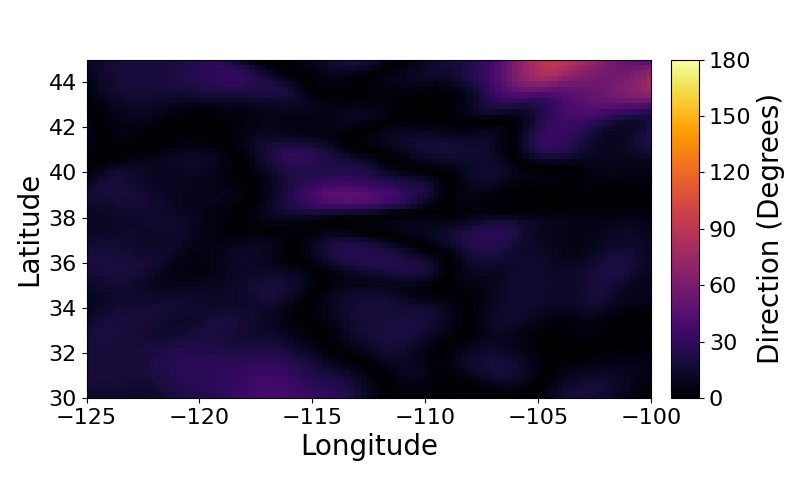}\label{fig:07_17_50_c}}
    
    \caption{ERA5 and Synthetic Model Variations on January 17 and July 17, 2023 at 000UTC at the 50 hPa Pressure Level}
    \label{fig:JanJulyVariation}
\end{figure*}

To generate realistic wind fields for higher vertical resolution and realistic deviations from ERA5 forecasts we generate synthetic wind fields based on aggregated and interpolated radiosonde data from a particular region. The international Radiosonde program is the largest source of high-resolution in-situ atmospheric meteorological measurements over 2000 radiosonde launch sites locations all over the world are launched twice a day. These radiosondes typically collect temperature, pressure, humidity, and wind velocity measurements from surface level to 25+ km; the data is then assimilated into weather forecasts. We collected our radiosonde data from the University of Wyoming's Upper Air Sounding Database \cite{UofWyUpperAir}. Throughout this work, we use the Southwestern United States region, shown in Figure \ref{fig:synthwinds}a for our station-keeping simulations. We have conducted several flight experiments of SHAB-Vs in this region and it contains an adequate number of radiosonde launch sites for generating synthetic forecasts. Figure \ref{fig:synthwinds}b shows all of the unprocessed wind velocity measurements recorded from radiosondes in the Southwestern United States region launched on August 23, 2023, at 1200 UTC. While radiosonde data has a very high vertical resolution compared to ERA5 forecasts,  they have poor horizontal and temporal resolution.  To account for this, we make several assumptions during aggregation and interpolation of the radiosonde data to generate synthetic forecasts.  

First, we bin the radiosonde data from an individual station into 250 m regions, taking the nearest $u$ and $v$ wind components.  Typically, there are multiple wind velocities per bin, in which case we take the closest value instead of an average (ex. with radiosonde altitude readings of 15998 and 16103 meters, we would select 15998 m for the 16000 m bin). In the case of empty bins, the bin is filled with $u$ and $v$ wind components via linear interpolation from the nearest filled bins. Next, the radiosonde data is interpolated horizontally using a nearest neighbor method; this generates coarse planar fields of winds at each altitude level for the area of interest. Finally, we apply Gaussian smoothing to each level to add smooth variation between the sub-areas of wind. Figure \ref{fig:synthwinds}c shows the final smoothed plane of a synthetic wind field at 16 km. 

Unfortunately, the synthetic winds are typically only available in 12-hour increments, at 0000 and 1200 UTC.  In simulation, we reduce this time gap from 12 hours to 3 hours to increase variation and ideally make the trained DRL HAB agents more robust to situations with highly dynamic forecasts. Three hours is also the temporal resolution of forward-predicted GFS forecasts. 
We hypothesize that if the trained DQN HAB agents are successful at station-keeping with large intraday wind shifts, they should also perform similarly or better on smoother changes.

\subsection{Comparing Synthetic Wind Forecasts with ERA5 Reanalysis Forecasts} \label{SynthForecastModeling_SynthCompareERA5}


When comparing the two forecasts, the synthetic winds forecasts are overall highly correlated to the ERA5 forecasts but also exhibit variance trends across certain pressure levels and seasons. Fig. \ref{fig:JanJulyVariation} shows the differences between the ERA5 and synthetic wind model for the SW USA region on January 17, 2023 and July 17, 2023. Fig. \ref{fig:01_17_50_c} shows the wind direction between an ERA5 and synthetic forecast at the 50 hPa pressure level on January 17, 2023, in the USA region. As shown visually, there is a significant difference in magnitude and direction with a mean angular difference of $61.6 \pm 47.2 \degree$ in January.  In this context, a large standard deviation is indicative of a non-uniform shift between the two models' wind direction. Similarly, a lower standard deviation may indicate that the model is inaccurate but has a uniform shift of some magnitude.
The same pressure level (50 hPa) on July 17, shown in Fig. \ref{fig:07_17_50_c} has a mean difference of $13.7 \pm 12.2 \degree$. When compared to the same region in January, the accuracy of the synthetic model in July is considerably better. 
This may be due to the increased turbulence in transition regions, and pressure levels between 50 hPa and 100 hPa, during the month of January. As a result, the number of inflection points can produce inaccuracies during the spatial interpolation. Subsequent sections analyze seasonal trends that help characterize these discrepancies.


\section{Discussion}\label{section:Discussion}

\subsection{Forecast Classification}\label{section:Discussion_ForecastClassification}

Station Keeping TWR success rates are completely dependent on the wind diversity for a region of interest, which can significantly vary seasonally and geographically.  To investigate how TWR success rates are affected by different regions and seasons we introduce a simple opposing winds binning algorithm to classify forecasts' wind diversity independently of trained HAB agents. The overall Forecast Score, $FS$, is a percentage of wind diversity based on the total count of opposing wind pairs in an altitude column at a specific central forecast subset coordinate, averaged across multiple time steps.


The wind bins are defined by:  

\begin{equation}
Bin_i = \Bigl\{(i-1)\cdot \frac{360\degree}{N_b}-\theta_{C}, 
(i)\cdot \frac{360\degree}{N_b}-\theta_{C},
\mathrm{...}
\Bigr\} , i = 1,2...,8
\end{equation}

where $N_b$ is the number of bins and $\theta_C$ is an angle offset center the cardinal directions within the bins (i.e. North (0\degree) is centered in Bin 1 with a range of [-22.5\degree,22.5\degree])

The Total opposing wind score for a specific coordinate and single timestamp is then defined by:

\begin{equation}
T_{\mathrm{opposing}} = \sum_{i=1}^{N_a}(C_i+C_{i+\frac{N_b}{2}})
\end{equation}

where $N_a$ is the number of altitude levels with binned wind values and $C_i$ is the total count in each respective wind bin. 

Finally, to get an overall forecast score, $FS$, with multiple timestamps for a particular coordinate,

\begin{equation}\label{forecastScoreEquation}
FS = \frac{T_{\mathrm{opposing},t_0} + T_{\mathrm{opposing},t_1} + \mathrm{...}}{n_t}\cdot 100
\end{equation}

where $n_t$ is the number of timestamps evaluated for opposing winds within a forecast subset. 

\begin{table}[h]
\renewcommand{\arraystretch}{1.3}
\caption{Percentage of Forecast Scores of Zero from 10,000 Samples}
\centering
\begin{tabular}{|c|c|c|}
\hline
\bfseries Month & \bfseries ERA5 Model & \bfseries Synthetic Model \\
\hline
January 2023 & 34.3\% & 19.2\% \\
April 2023 & 23.3\% & 6.49\%\\
July 2023 & 8.56\% & 0.18\%\\
October 2023 & 73.4\% & 38.1\%\\
\hline
\end{tabular}
\label{tab:zeroFSScores}
\end{table}

\begin{table*}[!t]
\renewcommand{\arraystretch}{1.3}
\caption{Mean ERA5 \& Synthetic Model Forecast Scores}
\centering
\begin{tabular}{|c|c|c|c|}
\hline
\bfseries Month & \bfseries ERA5 Model & \bfseries Synthetic Model  & \bfseries Score Difference\\
\hline
January 2023 & $0.388 \pm 0.21$ & $0.529 \pm 0.26$ & $0.141 \pm 0.17$ \\
April 2023 & $0.332 \pm 0.19$ & $0.632 \pm 0.23$ & $0.299 \pm 0.19$\\
July 2023 & $0.545 \pm 0.23$ & $0.736 \pm 0.18$ & $0.191 \pm 0.19$\\
October 2023 & $0.196 \pm 0.13$ & $0.399 \pm 0.23$ & $0.203 \pm 0.19$\\
\hline
\end{tabular}
\label{tab:FSDistributionMean}
\end{table*}

We sampled 10,000 randomized forecast subsets from the Southwest United States region for 4 different evaluation months in different seasons to compare forecast score distributions. The forecast scores were simultaneously computed for both the ERA5 and synthetic wind models at each randomized coordinate. First, we sampled the 10,000 randomized forecast subsets including impossible winds for station keeping (all winds blowing in one direction, forecast score of 0).
Table \ref{tab:zeroFSScores} lists the percentage of each month's forecasts that are zero scores from an unfiltered score distribution.  The percentage of zero scores in the month of October is staggering, nearly 73.4\% of unfiltered forecasts in the ERA5 model contain no opposing winds.

\begin{figure}[!htbp] 
    \centering
    \subfloat[ERA5 Filtered Forecast Score Distribution for SW USA]{\includegraphics[width=\linewidth]{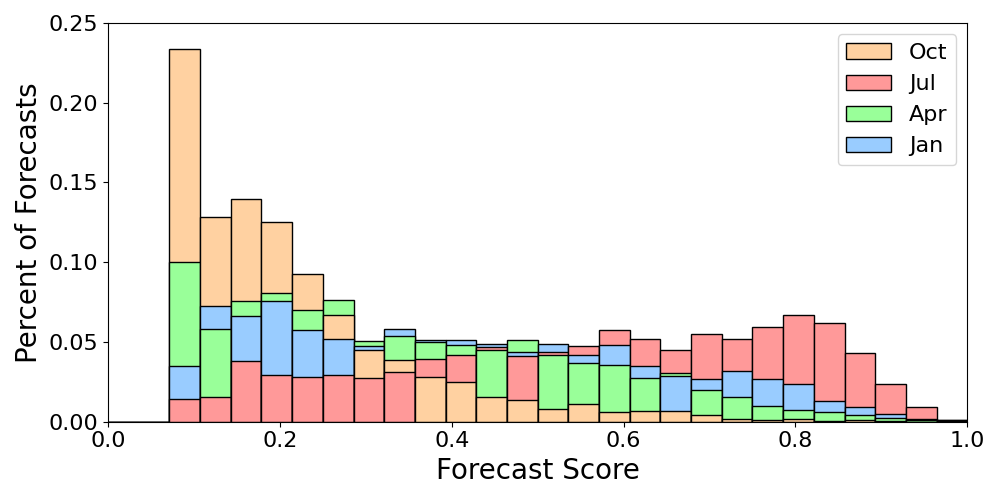}\label{fig:ERA5WindDistribution}}\\
    \subfloat[Synthetic Filtered Forecast Score Distribution for SW USA]{\includegraphics[width=\linewidth]{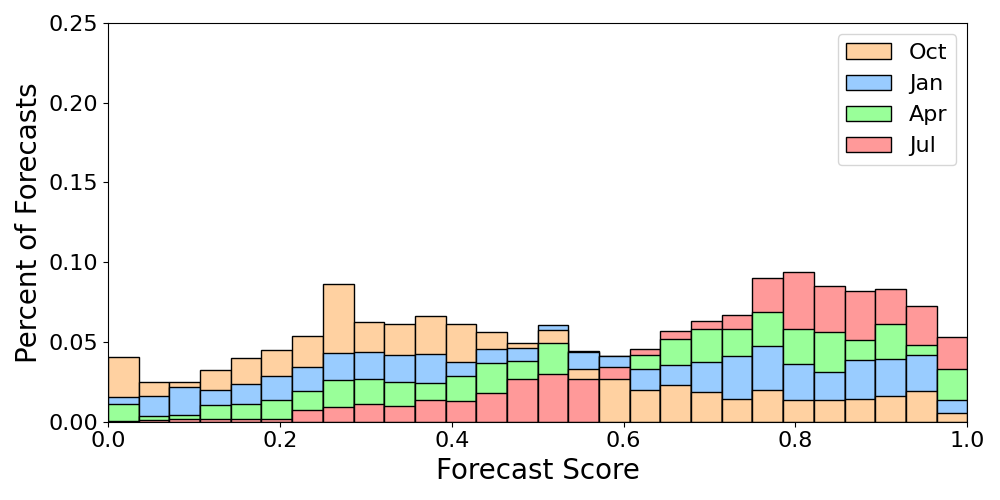}\label{fig:SynthWindDistribution}}
    
    \caption{Filtered Forecast Score Distributions for SW USA}
    \label{fig:windDistributions}
\end{figure}

Due to the inherent bias towards zero scores during the winter and fall months, we then recomputed the Forecast Score distributions, filtering out impossible winds (Forecast Scores \> 0), shown in Figure \ref{fig:ERA5WindDistribution} and Figure \ref{fig:SynthWindDistribution}.  It is important to note that the ERA5 Forecast Score distribution shown in Fig. \ref{fig:ERA5WindDistribution} does not contain values below 0.7 because the ERA5 forecast altitude region we use only contains 7 pressure levels. 

Table \ref{tab:FSDistributionMean} shows the mean and standard deviation of the filtered distributions found in Figures \ref{fig:ERA5WindDistribution} and \ref{fig:SynthWindDistribution} as well as the mean and standard deviation of the computed difference in forecast scores between the ERA5 and synthetic wind forecasts. These distributions suggest that it may be difficult to achieve station-keeping during the fall and winter months due to the low opposing wind probability. However, it is worth noting that these seasonal distributions may vary depending on the geographic region.

\begin{table}[b]
\renewcommand{\arraystretch}{1.3}
\caption{Overall Mean and Standard Deviation TWR50 Station Keeping Performance for four months in 2023}
\label{tab:overallTWR50}
\centering
\begin{tabular}{|c|c|c|c|c|}
\hline
\bfseries Month   & \bfseries Mean  & \bfseries SD & \bfseries Mean & \bfseries SD \\ 
        & \bfseries TWR50 & \bfseries TWR50   & \bfseries FS   & \bfseries FS \\
\hline
January &  20\%      &  0.25 & 0.38 & 0.21   \\
April   &  39\%      &  0.31 &  0.33 & 0.2  \\
July    &  51\%       &  0.31   & 0.54  & 0.23  \\
October &  31\%         & 0.23   & 0.19  & 0.13 \\ \hline
\end{tabular}%
\end{table}

For all four months selected, the ERA5 forecast score distribution underestimates the synthetic wind model distribution. This can be attributed to several factors, namely the lesser pressure level resolution. As mentioned in Section \ref{section:SimEnvforDQN}, the ERA5 reanalysis model on individual pressure levels operates at a vertical resolution of 7 pressure levels. Conversely, the synthetic wind model operates at a vertical resolution of 46 pressure levels. As a result, the forecast score computation allows for a naturally higher score for the synthetic wind model due to the availability of more in-between pressure levels.

As shown previously in Figures \ref{fig:01_17_50_c} and \ref{fig:07_17_50_c}, there can be significant variation in wind direction and magnitude at certain pressure levels which may also contribute to the non-uniform distribution of forecast scores. We conducted additional analysis to quantify the magnitude variation across different pressure levels in different seasons. These four figures are shown in the Appendix. 
January and July had relatively lower variance between ERA5 and synthetic forecasts than the seasonal transition months of April and October.
 These plots also consistently show that the pressure levels of 50 hPa and 70 hPa exhibit higher variation than other pressure levels.



\subsection{Forecast Score and TWR Evaluation}\label{section:Discussion_FS&TWR}




\begin{figure*}[!t]  
  \centering
  \begin{subfigure}[t]{0.25\textwidth}  
    \centering
    \includegraphics[width=\textwidth]{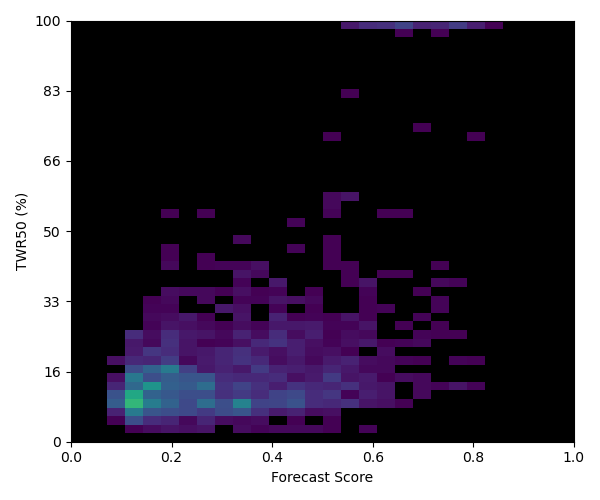}  
    \caption{January} \label{fig1}
  \end{subfigure}%
  \begin{subfigure}[t]{0.25\textwidth}  
    \centering
    \includegraphics[width=\textwidth]{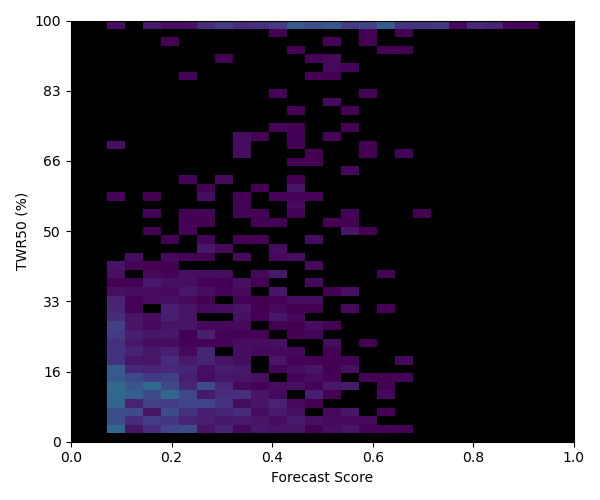}  
    \caption{April} \label{fig2}
  \end{subfigure}%
  \begin{subfigure}[t]{0.25\textwidth}  
    \centering
    \includegraphics[width=\textwidth]{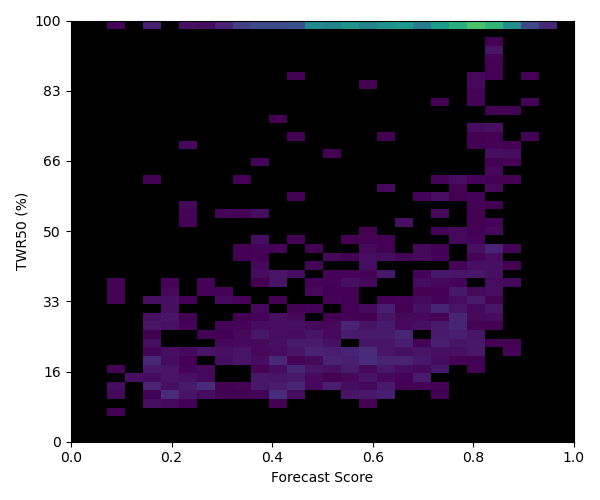}  
    \caption{July} \label{fig3}
  \end{subfigure}%
  \begin{subfigure}[t]{0.25\textwidth}  
    \centering
    \includegraphics[width=\textwidth]{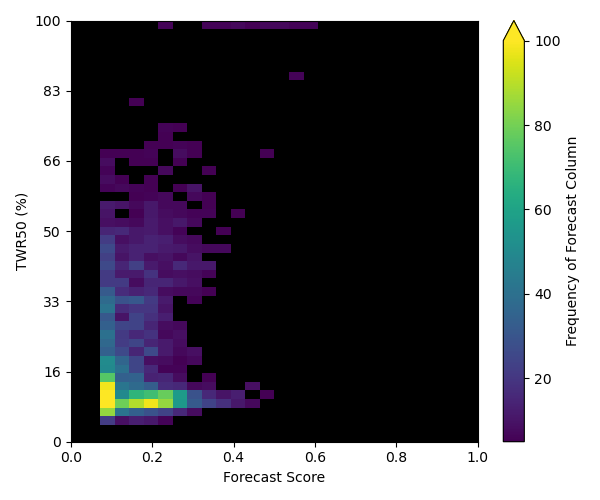}
    \caption{October} \label{fig4}
  \end{subfigure}

\caption{Frequency Distribution of Forecast Scores vs TWR50 evaluated with a trained DQN model on the months of January, April, July, and October 2023}
\label{fig:Heatmaps}
\end{figure*}


\begin{figure*}[!b]  
  \centering
  \begin{subfigure}[t]{0.25\textwidth}  
    \centering
    \includegraphics[width=\textwidth]{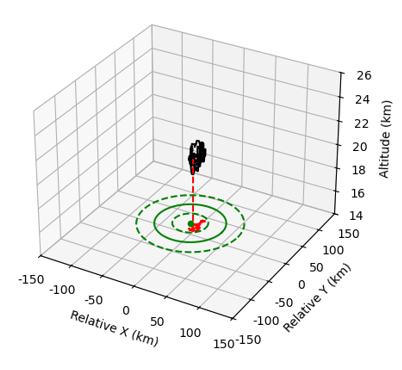}  
    \caption{January} \label{fig1}
  \end{subfigure}%
  \begin{subfigure}[t]{0.25\textwidth}  
    \centering
    \includegraphics[width=\textwidth]{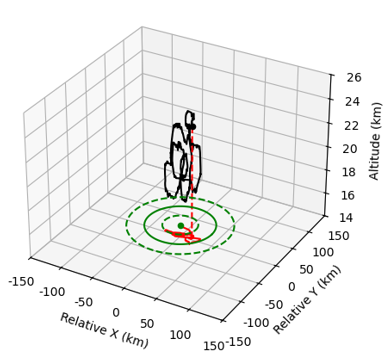}  
    \caption{April} \label{fig2}
  \end{subfigure}%
  \begin{subfigure}[t]{0.25\textwidth}  
    \centering
    \includegraphics[width=\textwidth]{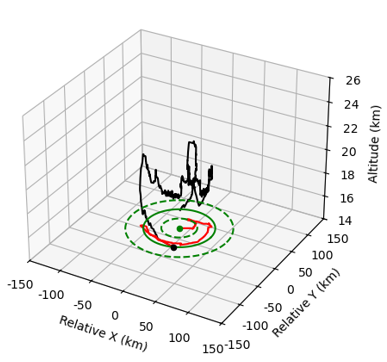}  
    \caption{July} \label{fig3}
  \end{subfigure}%
  \begin{subfigure}[t]{0.25\textwidth}  
    \centering
    \includegraphics[width=\textwidth]{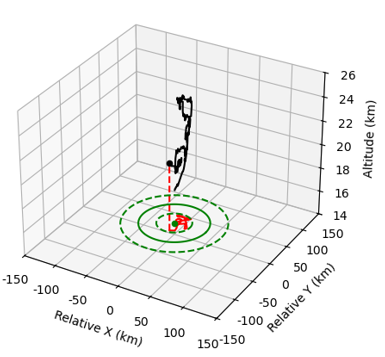}
    \caption{October} \label{fig4}
  \end{subfigure}

\caption{A Sampling of Station Keeping Trajectories with High TWR50s for each of the evaluation months in 2023}
\label{fig:Trajectories}
\end{figure*}

We evaluated our best-performing preliminary DQN model for 5,000 episodes each on 4 evaluation months in different seasons (January, April, July, and October). Table \ref{tab:overallTWR50} shows the overall station-keeping performance in each of the four evaluation months, with July having the highest Mean TWR50 as well as the highest Mean Forecast Score. Interestingly, April followed by October had the next highest Mean TWR50's at 20\% and 31\% respectively while having lower mean Forecast scores than January, which had the lowest Mean TWR50 of only 20\%. Overall, the standard deviations between the TWR50 and Forecast scores were similar across all four evaluation months, even though the different months have significantly different wind profiles. While Table \ref{tab:overallTWR50} shows a high-level summary between TWR50 and Forecast Score in different months, individual months have their own unique trends.

Figure \ref{fig:Heatmaps} shows initial TWR50 vs Forecast Score frequency distribution heatmaps from the same evaluation months on the same best-performing DQN model.  We filtered out the impossible winds for station keeping as discussed with the Forecast Score Distributions in the previous section, hence why each of the heatmaps have empty first two columns. We also filter out bins with less than 5 data points. The heatmaps forecast score distribution follows the aforementioned trends, where July has best wind diversity and October is dominated by poor winds. The occurrence of higher TWR50s also follows the same trend, with July having the most occurrences of perfect scores and October having the least. Because the Forecast Score is based on ERA5 forecasts, and not the synthetic winds, there are occurrences of perfect scores in all of the months.  Initial training on "perfect forecasts" using synthetic wind forecasts for both the wind column observation as well as the horizontal movement, suggests that significantly higher probabilities of perfect TWR50 scores (over 50\%) for high forecast scores (over 0.8) are possible with more forecast information. Figure \ref{fig:Trajectories} shows examples of unique station-keeping trajectories for HABs in each of the 4 evaluation months using the same trained DQN.

\section{Conclusions}

Altitude-controlled HABs such as SHAB-Vs can leverage opposing winds at various altitude levels to conduct short-duration station-keeping missions; the winds are constantly changing, and frequently unknown, making path planning very difficult. Recently, deep reinforcement learning has become a popular tool in robotics for path planning and obstacle avoidance in complex dynamic environments.  In this work, we developed a custom simulation environment for training and evaluating the station-keeping performance of short-duration HABs with deep reinforcement learning using Deep Q-Networks (DQN). A major limitation of training HABs for station keeping in simulation and transitioning to real-world flights is a lack of high-resolution wind data to include.  ERA5 reanalysis forecasts have historical hourly wind data at specific pressure levels from 1940 to the present but significantly lack vertical resolution, with typically less than 10 pressure levels included in the maneuverable altitude region, with up to several kilometer gaps between levels.  

We introduced a new strategy for generating synthetic wind forecasts from historical radiosonde data to create realistic deviations from the ERA5 forecast. From our initial analysis of forecasts in the Southwestern United States for several months in 2023, we show that overall, the synthetic wind forecasts tend to have a high correlation with ERA5 forecasts. Because the Synthetic Winds are aggregated from real balloon data, this leads us to assume that synthetic winds at altitude levels in between the mandatory pressure levels from ERA5 forecasts are realistic to real-world.  The biggest limitation with using Radiosondes for generating synthetic forecasts is the spare launch sides worldwide, as well as a lack of temporal resolution, with radiosondes typically only being launched twice a day.  

With DQN, we successfully trained short-duration HABs in simulation to station-keep and maintain time within a 50km region (TWR50) approximately 50\% of the time.  The best models for station-keeping short-duration HABS are highly dependent on when and where the HABs are launched.  The best months for station keeping have both high wind diversity and high wind velocity correlations between the synthetic forecasts and ERA5 forecasts.  To help predict station-keeping performance based on an ERA5 forecast, we introduced a forecast classification algorithm independent of the DQN algorithm that directionally bins winds at each altitude level and totals how many levels have opposing wind pairs, with higher totals leading to higher Forecast Scores.  Overall, The Forecast Score classification method shows higher Forecast Scores (typically over 60\%) have higher probabilities of successful TWR50 station-keeping performance.   Monthly evaluation of trained agents compared with Forecast Scores shows that probabilities of station-keeping success with high Forecast Scores vary, with April and July having higher probabilities than January and October. 

\subsection{Future Work}

In the future, we plan to explore more advanced interpolation, noise application, and smoothing strategies to increase the temporal resolution of the synthetic winds while remaining coupled to changes in the known ERA5 reanalysis forecasts. We also plan to compare synthetic forecasts and the performance of trained DQN agents on GFS forecasts as opposed to ERA5 forecasts, which are forward predicted, have lower temporal resolution than ERA5 forecasts, and most likely have lower correlation.  We plan to integrate the trained DQN algorithms on SHAB-Vs to evaluate station-keeping performance on real short-duration HABs.  Before conducting flight tests, we want to continue training agents on larger subsets of forecast data, instead of monthly forecasts, to be more robust to winds with high variance from seasonal trends and other unforeseen weather conditions.  To help transition from simulated agents to HABs, we also plan to develop an indoor testbed with miniature autonomous blimps and artificial wind fields using fans to evaluate and iterate our algorithms on easy-to-deploy lighter-than-air platforms.




\bibliographystyle{IEEEtran}
\bibliography{main}

\thebiography

\begin{biographywithpic}
{Tristan Schuler}{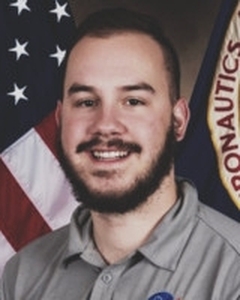}
is a Robotics Research Scientist at the U.S. Naval Research Laboratory (NRL) in the Distributed Autonomous Systems Section for the Navy Center for Applied Research in Artificial Intelligence (NCARAI). He
is an expert in autonomous lighter-than-air platforms with several publications and patents since 2018. Tristan has a B.S. in Mechanical Engineering from George Mason University and an M.S. in Aerospace Engineering from the University of Arizona where he defended his thesis on solar balloons for planetary exploration. In 2021, he was awarded a Karles research Fellowship from the NRL to develop autonomous HAB technology for SHAB-Vs. His HAB expertise has led to continued research and collaborations with AFRL, NRL, ONR, Sandia, NASA JPL, and several universities.
\end{biographywithpic} 

\begin{biographywithpic}
{Chinthan Prasad}{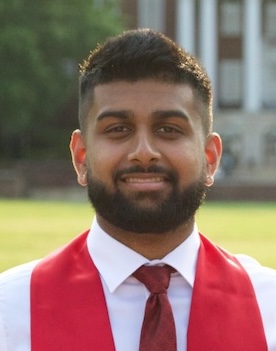}
is a Research Scientist at the U.S. Naval Research Laboratory (NRL) with experience in lighter-than-air-vehicles, 18-DOF legged robots, and simulator development. He is a full-time graduate student at the University of Maryland pursuing a Master of Engineering degree in Aerospace Engineering with a concentration in Robotics and Machine Learning. He also achieved his B.S. in Aerospace Engineering from the University of Maryland with a minor in Computer Science. As a member of the Distributed Autonomous Systems Section of the Navy Center for Applied Research in Artificial Intelligence (NCARAI), he has contributed to multiple research efforts on Robotic Gait Optimization, Swarming, and Navigation. 
\end{biographywithpic}

\begin{biographywithpic}
{Georgiy Kiselev}{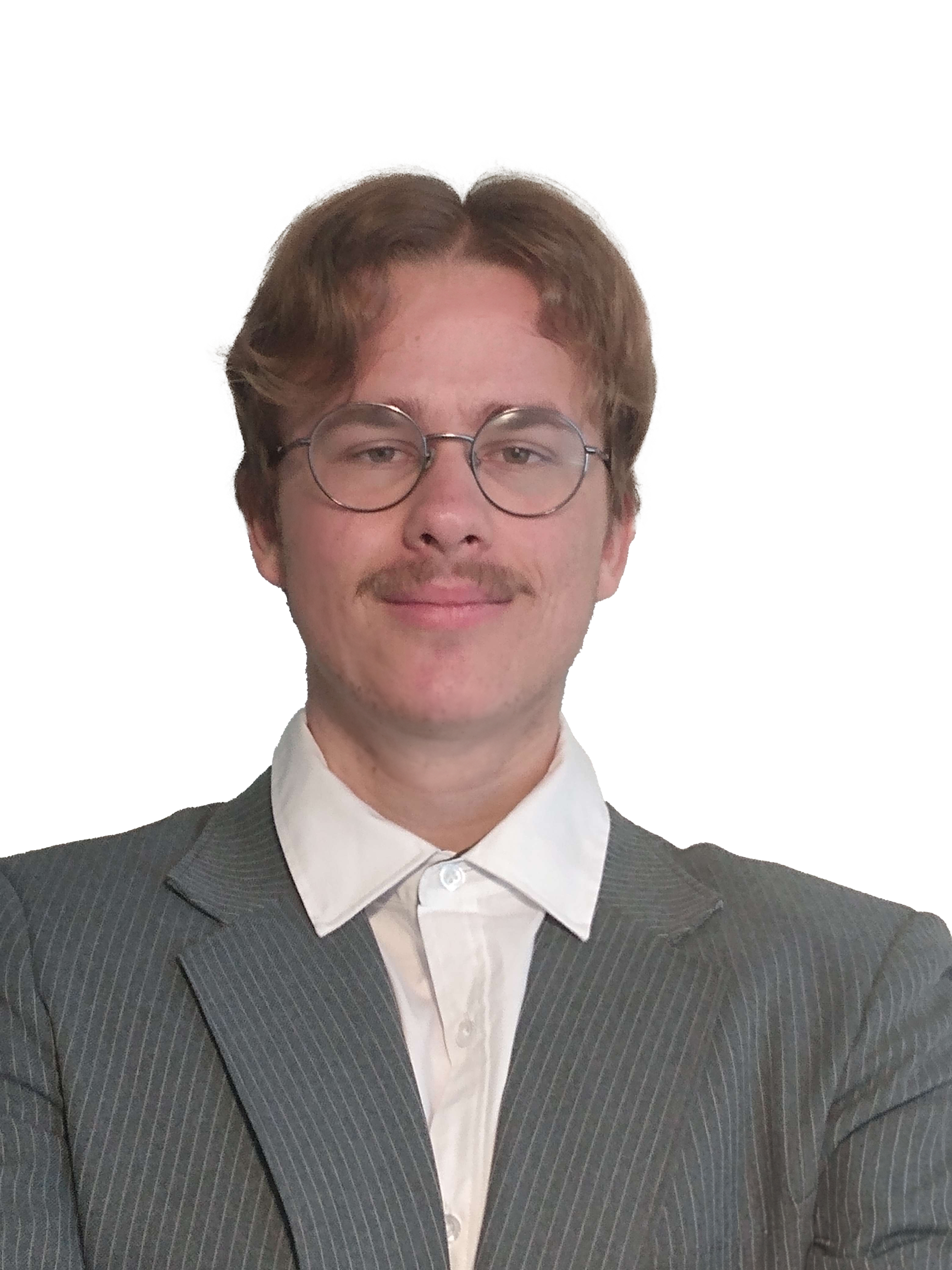}
is an undergraduate student at the University of California Davis pursuing a B.S. in Statistics with a focus on Machine Learning. He specializes in biomedical image processing and has created multiple machine learning diagnosis software. In 2023, he received several awards from campus based organizations such as HackDavis and GDSC for a variety of biomedical machine learning applications like a Chest Radiology Application and Retinal Vascular Tree Extractor. He has also contributed his expertise in Computer Vision towards the development of software for the diagnosis of various rice plant diseases. 
\end{biographywithpic}

\begin{biographywithpic}
{Donald Sofge}{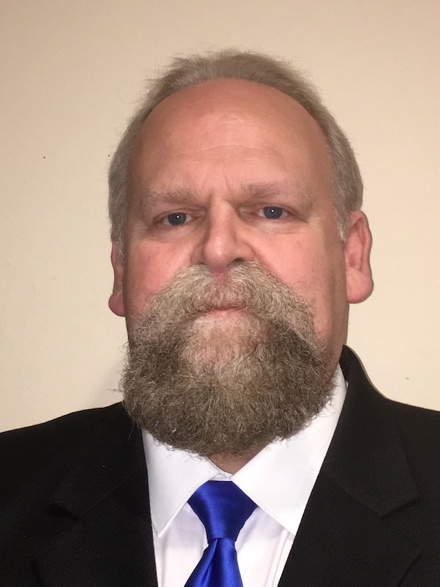}
Don Sofge is a Computer Scientist and Roboticist at the Naval Research Laboratory (NRL) with 36 years of experience (23 at NRL) in Artificial Intelligence, Machine Learning, and Control Systems R\&D. He leads the Distributed Autonomous Systems Section in the Navy Center for Applied Research in Artificial Intelligence (NCARAI), where he develops nature-inspired computing paradigms to challenging problems in sensing, artificial intelligence, and control of autonomous robotic systems. He has served as PI/Co-PI on dozens of federally-funded R\&D efforts, and has more than 200 publications (including 12 books) in robotics, artificial intelligence, machine learning, planning, sensing, control, and related disciplines. He has served as an advisor on autonomous systems to DARPA, ONR, OSD, ARL, NSF, and NASA, as well as US representative on international TTCP and NATO technical panels on autonomous systems.
\end{biographywithpic} 
\newpage
\onecolumn
\section*{Appendix}
\addcontentsline{toc}{section}{Appendix}

\begin{figure*}[!ht] 
    \centering
    \begin{subfigure}[h]{\textwidth}
        \centering
        \includegraphics[width=\textwidth]{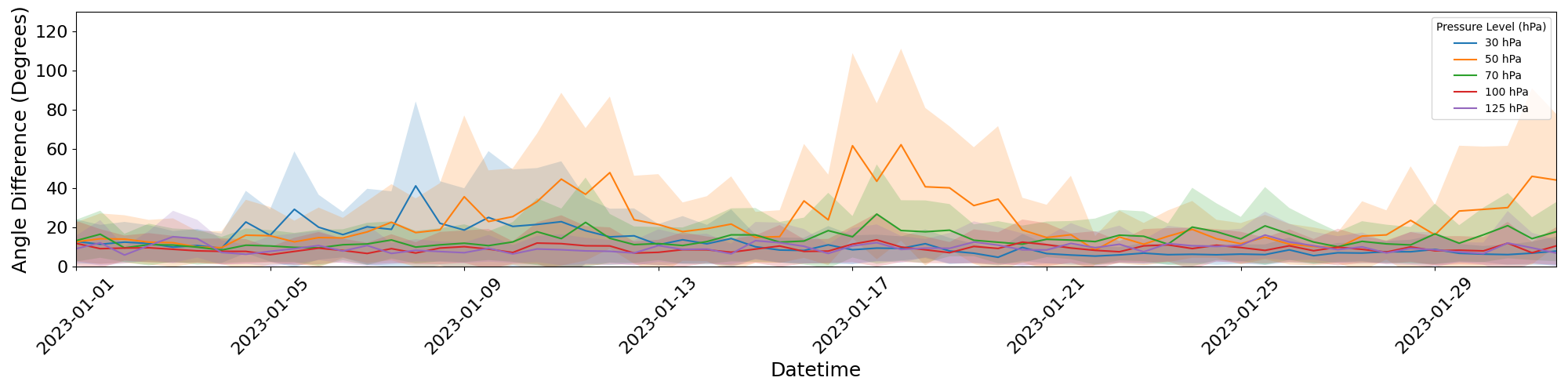}
        \caption{Mean and Standard Deviation of Model Variation in January 2023}
        \label{fig:ModelVariationJanuary}
    \end{subfigure}
    
    \vspace{4mm} 

    \begin{subfigure}[h]{\textwidth}
        \centering
        \includegraphics[width=\textwidth]{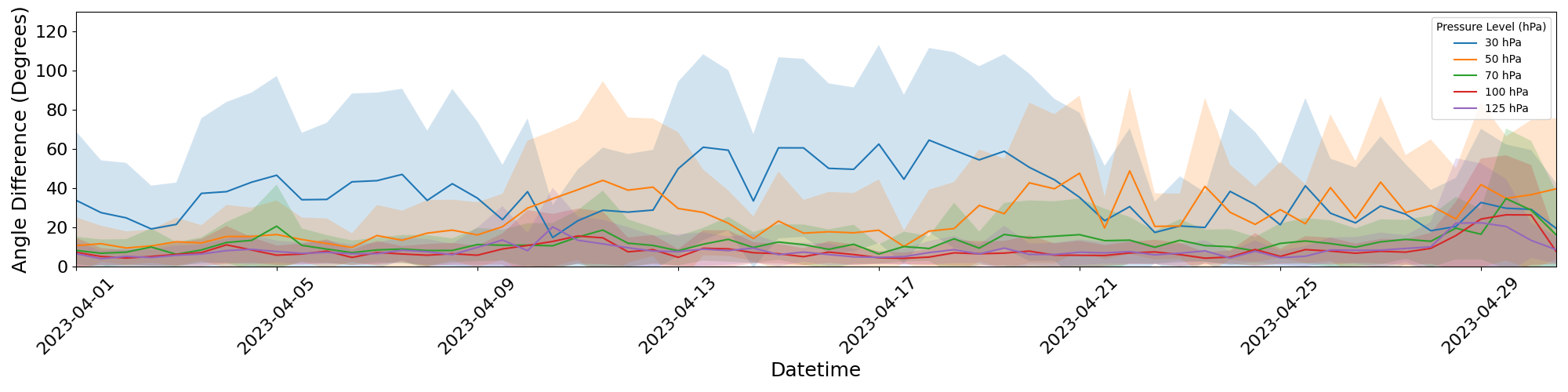}
        \caption{Mean and Standard Deviation of Model Variation in April 2023}
        \label{fig:ModelVariationApril}
    \end{subfigure}
    
    \vspace{4mm} 

    \begin{subfigure}[h]{\textwidth}
        \centering
        \includegraphics[width=\textwidth]{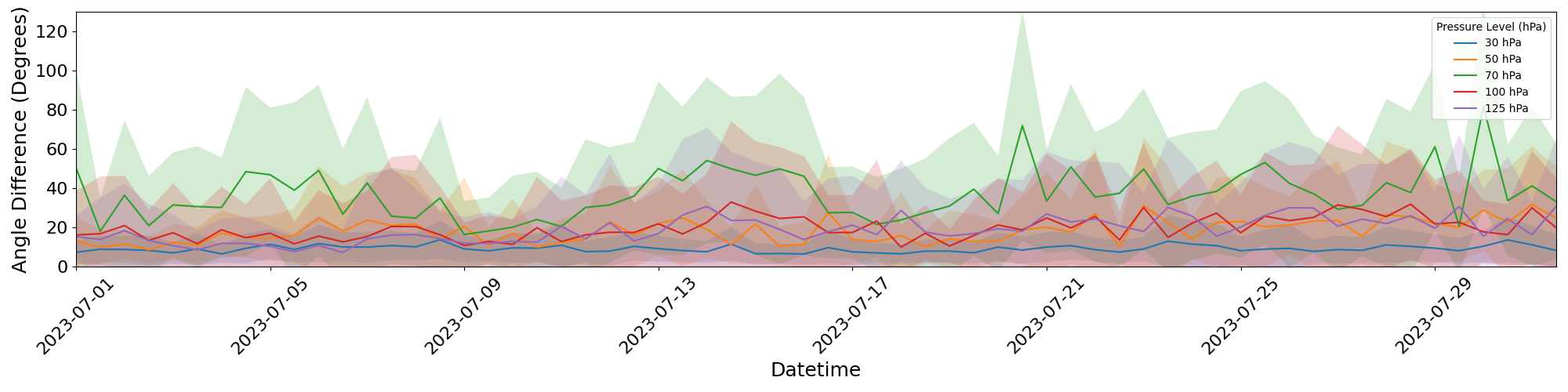}
        \caption{Mean and Standard Deviation of Model Variation in July 2023}
        \label{fig:ModelVariationJuly}
    \end{subfigure}
    
    \vspace{4mm} 

    \begin{subfigure}[h]{\textwidth}
        \centering
        \includegraphics[width=\textwidth]{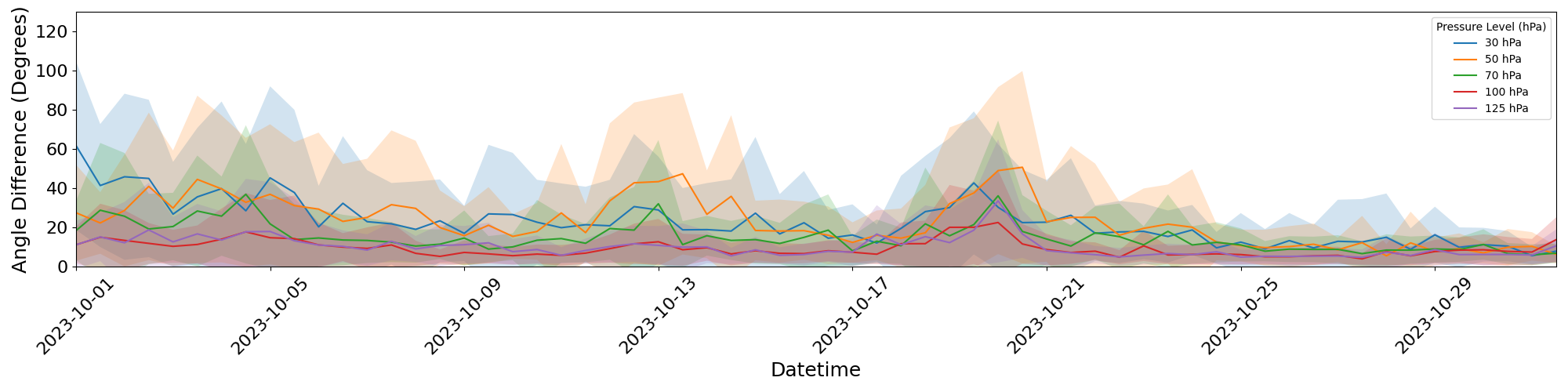}
        \caption{Mean and Standard Deviation of Model Variation in October 2023}
        \label{fig:ModelVariationOctober}
    \end{subfigure}

    \caption{Mean \& Standard Deviation of Model Variation at Different Pressure Levels in Southwestern USA}
    \label{fig:ModelVariations}
\end{figure*}

\end{document}